\documentclass[conference]{IEEEtran}
\IEEEoverridecommandlockouts
\usepackage{cite}
\usepackage{amsmath,amssymb,amsfonts}
\usepackage{algorithmic}
\usepackage[colorlinks=true, citecolor=blue, linkcolor=black]{hyperref}
\usepackage[nameinlink,noabbrev]{cleveref}
\usepackage{graphicx, caption, subcaption}
\usepackage{textcomp}
\usepackage{xcolor}
\usepackage{caption}
\usepackage{pifont}
\newcommand{\xmark}{\ding{55}}%
\def\BibTeX{{\rm B\kern-.05em{\sc i\kern-.025em b}\kern-.08em
    T\kern-.1667em\lower.7ex\hbox{E}\kern-.125emX}}
\begin{document}

\title{Enhancing E-Commerce Recommendation using Pre-Trained Language Model and Fine-Tuning \\
\thanks{This work was done in 2021.}
}

\author{\IEEEauthorblockN{Nuofan Xu}
billxusuzhou@gmail.com
\and
\IEEEauthorblockN{Chenhui Hu}
chenhui.hu@gmail.com
}

\maketitle

\begin{abstract}
Pretrained Language Models (PLM) have been greatly successful on a board range of natural language processing (NLP) tasks. However, it has just started being applied to the domain of recommendation systems. Traditional recommendation algorithms failed to incorporate the rich textual information in e-commerce datasets, which hinderss the performance of those models. We present a thorough investigation on the effect of various strategy of incorporating PLMs into traditional recommender algorithms on one of the e-commerce datasets, and we compare the results with vanilla recommender baseline models. We show that the application of PLMs and domain specific fine-tuning lead to an increase on the predictive capability of combined models. These results accentuate the importance of utilizing textual information in the context of e-commerce, and provides insight on how to better apply PLMs alongside traditional recommender system algorithms. The code used in this paper is available on Github: \textcolor{blue}{ https://github.com/NuofanXu/bert\_retail\_recommender}.
\end{abstract}

\section{Introduction}
With the rapid development in web technology, Internet users have been provided with tremendous amount of information, which impedes the selection of relevant information to the user, resulting in protracted decision-making process and reduced decision quality. To address this problem of information overload, solid efforts have been made by researchers in various domains to develop recommendation systems that are able to generate curated information based on user preferences, ranging from online advertisement\cite{Chaudhuri2016}, music selection\cite{Kowald2020}, move rating\cite{Nguyen2013}, to product recommendation\cite{Agarwal2018}. By exploiting user-item (UI) relations, recommender systems recommender systems hold the promise of streamlining data and provide personalization, and have been a key determining factor for driving a nuanced customer experience\cite{Ricci2015}. As a result, they have been widely adopted by online retailers and web service providers to perfect their products and acquire profit. \\\\
E-commerce has gained increasing importance and popularity among customers. Although recommendation systems are now intrinsic to many industries, there are some unique challenges that thwart the widespread adoption in the field of E-commerce. In most of the cases, potential customers do not have a crisp understanding of what they want to buy. It is hard for them to accurately express their need with limited taxonomy of attributes with the internal search engine.\cite{Agarwal2018} As a result, they get discouraged fairly easily in the process of navigating through a plethora of options. Undoubtedly, the capability of recommender system implemented by those online retailers to capture the intrinsic item characteristics and user preferences has becoming one of the biggest differentiators against their competitors and is critical to ensuring the success of a company\cite{Ricci2015}.\\\\
There is largely a dichotomy of recommendation algorithms. Content-based algorithms build user specific and item specific profiles, usually in the form of a vector. The user profile describes the contents that he or she likes, and the item profile describes the contents that it contains. To create those profiles, a good amount of  auxiliary information of users and items are required. Then a similarity metric, typically a dot product, is used to determined user preferences. Collaborative filtering approach analyzes past user behavior and finds similar users and items based on the history. The core assumption here is that the users who have agreed in the past tend to also agree in the future. People have also been using deep learning (DL) techniques to further improve the performance of those recommender models, as DL provides better modelling power, flexibility, and feature representation learning. Covington et al.\cite{Covington} presented a deep neural network based recommender for video recommendation on YouTube. Cheng et al. \cite{Cheng2016} proposed a wide \& deep model for generic recommender systems with sparse inputs. Shumpei et al. \cite{Okura2017} showcased a recurrent neural network (RNN) based news recommender system for Yahoo News. All of these deep learning based models have proven to have significant improvement over traditional models. Hybrid model refers to recommender system that integrates two or more types of recommendation methods.

While most of the existing recommendation algorithms are powerful in modeling user-item relationships, they have major limitations regarding the utilization of text information. Specifically, they cannot effectively incorporate textual information in e-commerce datasets, which are in abundance and might potentially be very informative. Without preparations, those recommendation algorithms also suffer acute relevance problems due to lack of negative feedback in the context of online retail\cite{Jiang2020}. The problem regarding which recommender algorithms to use and how to effectively use those algorithms, while heatedly debated and investigated, remains challenging and unsolved.\\\\
The recent emergence and development of BERT (Bidirectional Encoder Representations from Transformers) and its variants in NLP field has drastically increased the modeling power of machine learning models to understand textual information\cite{Devlin2019}. By leveraging the power of pretraining contextual representation on large scale text datasets and unsupervised training objectives, BERT has been able to achieve state-of-art performances in a board range of natural language processing tasks. \\\\
However, in the field of e-commerce, people could not exploit the full advantage of BERT algorithms due to the following reasons: BERT cannot always accurately understand the frequently appearing domain specific phrases because they are not included in the training objective; in the context of e-commerce, textual information such as item label or item description typically has very limited length compared with other data sources such as movie reviews or web documents, on which BERT is hard to make inference without additional domain specific pretraining and is very susceptible to noises in the dataset\cite{Zhang2020}. \\\\
It remains unclear that how the benefit of a unified language representation model would help the recommendation models to better understand and capture the underlying rules that affects people's decision making, and how different schemes of further pretraining would influence the final recommendation results. In our paper, we address this question for a high performing pre-trained language model, RoBERTa\cite{Liu2019}, and investigate the effectiveness of various approaches to incorporate rich text features into traditional recommender system algorithms on an online retail data set. We consider two types of recommendation algorithms - content-based recommender algorithm (matrix factorization, abbreviated as MF) and machine learning based algorithm (XGBoost and decision tree). We evaluate the performance increase gained through the incorporation of sentence embeddings against baseline algorithms, and we show that the introduction of RoBERTa consistently increases model performance regardless of which recommender algorithm is used. Additionally, the effect of pretraining BERT with transformer-based denoising auto-encoder (TSDAE) and masked language model(MLM) is also investigated. For MLM, in accordance to the RoBERTa paper, we use dynamic masking rather than static masking\cite{Liu2019}. Inspired by this process, We further propose a recommendation scheme that we consider to be the most effective for MF and XGBoost respectively. Our experiment shows that recommendation models trained under this scheme outperforms their vanilla counterparts even with highly sparse user-item interaction matrices. \\

In summary, our contribution includes:
\begin{itemize}
    \item a comprehensive study on incorporating rich textual information in the domain of e-commerce to improve performance of recommender algorithms;
    \item a thorough evaluation and visualization of recommendation results made by vanilla recommendation algorithms and BERT enhanced recommendation models;
    \item an investigation into the effectiveness of domain-adaptive pretraining.
\end{itemize}

\section{Related Work}
\subsubsection{Matrix Factorization}
A straightforward matrix factorization model maps both users and items to a joint latent factor space of dimensionality K — such that user-item interactions are modeled as inner products of user feature vector and item feature vector in that space\cite{Nguyen2013}. However, pure matrix factorization relies on past information, and cannot accurately capture some observed variation in the rating values are due to biases associated with either users or items. Xxx solves this by using a new training objective and error term with additional bias term\cite{Tran2018}.  However, the cold start problem remains challenging due to its inability to address new items and new users\cite{Elahi2016}. Wang et al. Integrates additional user attributes (demographics)  to relieve this problem\cite{Wang2012}, though the performance of matrix factorization model still suffers from high sparsity of the user-item interaction matrix. It is also common in many real-world use cases that we only have access to implicit feedback, where user-item interactions like clicks, likes, or purchases are used as a proxy to indicate a preference or dislike for those items. Standard matrix factorization-based collaborative filtering treats the entries of the user-item matrix as explicit user preferences given to items. Those algorithms assume explicit data, where the user has rated both things they like and dislike using some rating scale. They typically work by treating the missing data as an unknown, set them to some constant (usually 0), and then the decomposed matrices are trained to reproduce the entire matrix, including the unknown values using SGD. Hu et al. proposed implicit feedback\cite{Hu2008}. Essentially, instead of trying to model the matrix of ratings directly, implicit matrix factorization is used to treat the data as numbers representing the strength in observations of user actions (such as the number of clicks, or the cumulative duration someone spent viewing a movie). Those numbers are then related to the level of confidence in observed user preferences, rather than explicit ratings given to items. The model then tries to find latent factors that can be used to predict the expected preference of a user for an item.

\subsubsection{Pre-trained Language Model}
Recently, substantial work has shown that pre-trained language models (PLMs) can learn universal language representations\cite{Peters2018, Radford2018, Devlin2019}. Those language representations learned on large corpus avoid training a new model from scratch and allow faster and better implementation of NLP tasks.
Compared with traditional word embedding models\cite{Gupta2015}, PLMs learn to predict and represent words based on the entire input context, thus are capable of understanding semantics more accurately. Although PLMs have been successful on a broad range of NLP task, there are still lots of areas of improvement to be made. Many efforts have been done to further optimize its robustness and model performance. SpanBERT\cite{Joshi2020} proposes random masked span reconstruction instead of single word reconstruction. However, the span consists of random continuous words and may not form phrases, thus fails to capture phrase-level knowledge accurately. ERNIE-1.0\cite{Sun2019} integrates phrase-level masking and entity-level masking into BERT. RoBERTa\cite{Liu2019} proposes modifications to the BERT pretraining procedure that improve end-task performance, allowing PLMs to further untap its potential. 
PLMs have proved its ability on sentence completion tasks (including modest performance in Winograd challenge sentences) if trained on a large corpus. 
\subsubsection{XGBoost}
Among many powerful machine learning algorithms, gradient tree boosting is one of the methods that has been proven to shine in many standard classification benchmarks. XGBoost is a fast and highly scalable implementation for tree boosting. It handles sparse input with a novel tree learning algorithm and handles instance weights
in approximate tree learning with
As recommendation systems are being used for a long time, several algorithms for recommendation have been developed. Literature shows that extending model based algorithms \cite{Martin2017} is the most promising class of algorithms. In this class, several machine learning techniques are adapted to work with user-item interactions as well as interaction data, and user and item properties. One of those techniques which appeared to be very successful in earlier studies is decision tree learning. In this method a decision tree is used to predict the target value of an item by means of observation that have been made. By using the gradient boosting technique, models are added sequentially to correct the errors of prior models until no further improvements can be made.

\subsubsection{Fine-tuning PLMs}
Fine-tuning a pre-trained language model (PLM) has become the standard for doing transfer learning in natural language processing. Over the last three years, fine-tuning\cite{Howard2018} has superseded the use of feature extraction of pre-trained embeddings\cite{Peters2018} while PLMs are favoured over models trained on translation \cite{McCann2018}, natural language inference \cite{Conneau2017}, and other tasks due to their increased sample efficiency and performance \cite{Zhang2019b}. The empirical success of these methods has led to the development of even larger models \cite{Devlin2019, Raffel2020}. Recent models are so large in fact that they can achieve reasonable performance without any parameter updates\cite{Brown2020}. The limitations of this zero-shot setting, however, make it likely that in order to achieve the best performance or stay reasonably efficient, fine-tuning will continue to be the modus operandi when using large PLMs in practice.

\section{Datasets and Data Preparation}
Three different text datasets are used in this paper. We use Online Retail dataset\cite{Chen2019} as our main target of study. It focuses on a narrow domain of e-commerce, as this British Online retailer sells a limited number of things during the period when this dataset is collected. We use Amazon dataset as a dataset that is similar but covers a broader range of things, as they are both e-commerce datasets. The openwebtext dataset\cite{Cohen2019} is considered as a close analogy to what text datasets are used for RoBERTa training, and it is only used for cross reference when computing vocabulary similarity. Only part of it is sampled due to its much larger size compared to the other two datasets.\\

\begin{figure*}
	\centering
	\includegraphics[scale=0.6,angle=90]{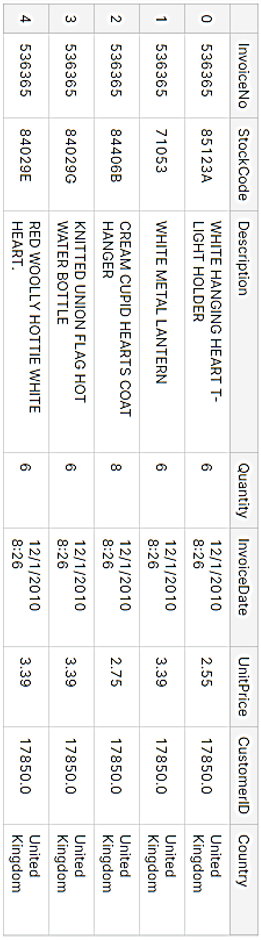}
	\caption[snippet of Online Retail dataset]
	{a snippet of the Online Retail dataset}
	\label{tab:figure0}
\end{figure*}

\begin{figure}
	\centering
	\includegraphics[scale=0.8]{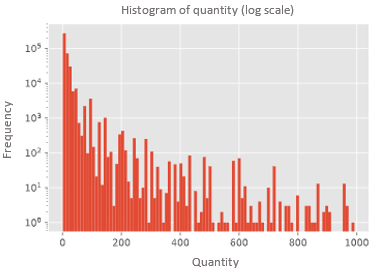}
	\caption[Close up of]
	{histogram of purchased quantity of items in online retail dataset}
	\label{tab:figure1}
\end{figure}

\begin{figure}
	\centering
	\includegraphics[scale=0.7]{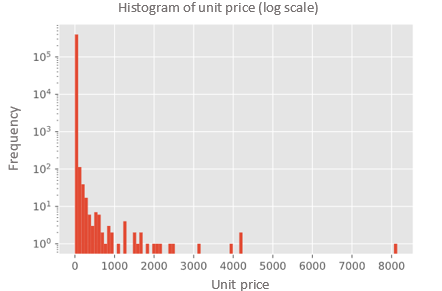}
	\caption[Close up of]
	{histogram of unit price of items in online retail dataset}
	\label{tab:figure2}
\end{figure}

The online retail dataset consists of transactions occurring between 01/12/2010 and 09/12/2011 for a UK-based and registered non-store online retail. The company mainly sells unique all-occasion gifts. Many customers of the company are wholesalers. 
Each observation has eight covariates, which are invoice ID, stock code, item description, quantity, date, unit price, customer ID, and country. The invoice ID is a 6-digit number uniquely assigned by the retail to be able to identify if different items belong to the same user. Canceled transactions is indicated by invoice number which starts with letter 'C'. The stock code is a 5-digit integral number uniquely assigned to each distinct product, while the customer ID is a 5-digit user specific integer. The item description is a short sentence describing what the item is. Date is the day and time when each transaction was generated. The last one is country, showing the name of the country where each customer resides. There are 37 countries in total, ranging from North America to EU, with most customers from UK and France. The first five lines of the dataset is shown in \autoref{tab:figure0}.
\\
There are some special entries and interesting characteristics in the dataset that are worth mentioning:
\begin{itemize}
    \item Transactions with negative quantity 
    \begin{itemize}
        \item The magnitude of those negative values match with previous purchase records in the dataset. 
        \item In the context of online retail, it means some items got returned.
    \end{itemize}
    \item Transactions with zero or negative price. 
    \begin{itemize}
    \item Transactions with zero price does not bring useful information to analysis, moreover mainly there are not any description for such transactions. 
    \item Transactions with negative price. From the description we could see that these transactions were probably made as corrections.
    \end{itemize}
    \item 	Long-tailed data distribution with extreme outliners   
    \begin{itemize}
        \item 	Right-skewness of quantity and unit price. As shown in \autoref{tab:figure1}, even in log scale, the data is still right-skewed. In fact, 50\% of the users only purchase 1-3 items at a time, while some wholesaler does big purchases with more than 500 of the same items. In addition, most items sold in the dataset are cheap, with rare cases of high unit prices. As a result, those data are extremely right-skewed.
        \item Outliners. From \autoref{tab:figure2}, most of items are common, daily products which have a normal price. However, some items in this dataset are very expensive, and that might influence the amount of purchase significantly. Also, standard deviation for quantity is 218.08, which is extremely huge. This is caused by some rare and extreme events. For example, a big purchase of 80955 items happened during the period when the dataset is recorded, but unfortunately this order is eventually cancelled and thus is not taken as part of our data.\\   
    \end{itemize}
\end{itemize}

There are 1336 of transactions with negative quantity value among non-cancelled transactions, and 1336 transactions with zero price and negative quantity. This corresponds to 1.71\% and 0.46\% of the total number of observation entries in the dataset, which is insignificant and less likely to cause loss of information if those problematic entries are directly filtered out. 
There are 541909 entries in this online retail dataset, and there are 406829 non-null customer IDs and 540455 item descriptions, meaning 24.9\% of customer IDs and 0.27\% of item descriptions are null object and thus cannot be used. After filtering out all the null entries and return cases, there are 397924 observations. Those records correspond to 4339 distinct users and 3665 unique items. This gives a user-item matrix with a sparsity of 97.5\%. 

\begin{figure}
	\centering
	\includegraphics[scale=0.8]{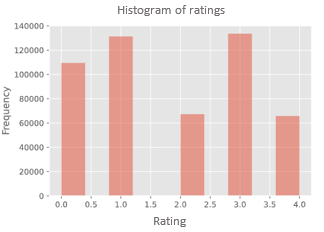}
	\caption[Close up of]
	{histogram of ratings after quantity-to-rating conversion in online retail dataset}
	\label{tab:figure3}
\end{figure}

To reduce the influence of outliners and uneven distribution on the performance of our models, there are normally two methods. We can standardize the features by subtracting mean and dividing the result by standard deviation, but that would result in negative values in features and potentially negative output values in matrix factorization. Here we choose the second method, evening out the distribution by creating a histogram as shown by \autoref{tab:figure3} and assign labels to each quantity range that approximately contains the same amount of products. Due to the discrete nature of this method, each label would not have the same amount of items, but as shown in \autoref{tab:figure3}, the distribution is much better than the right-skewed one shown in \autoref{tab:figure1}.

\begin{figure*}
	\centering
	\includegraphics[scale=0.45]{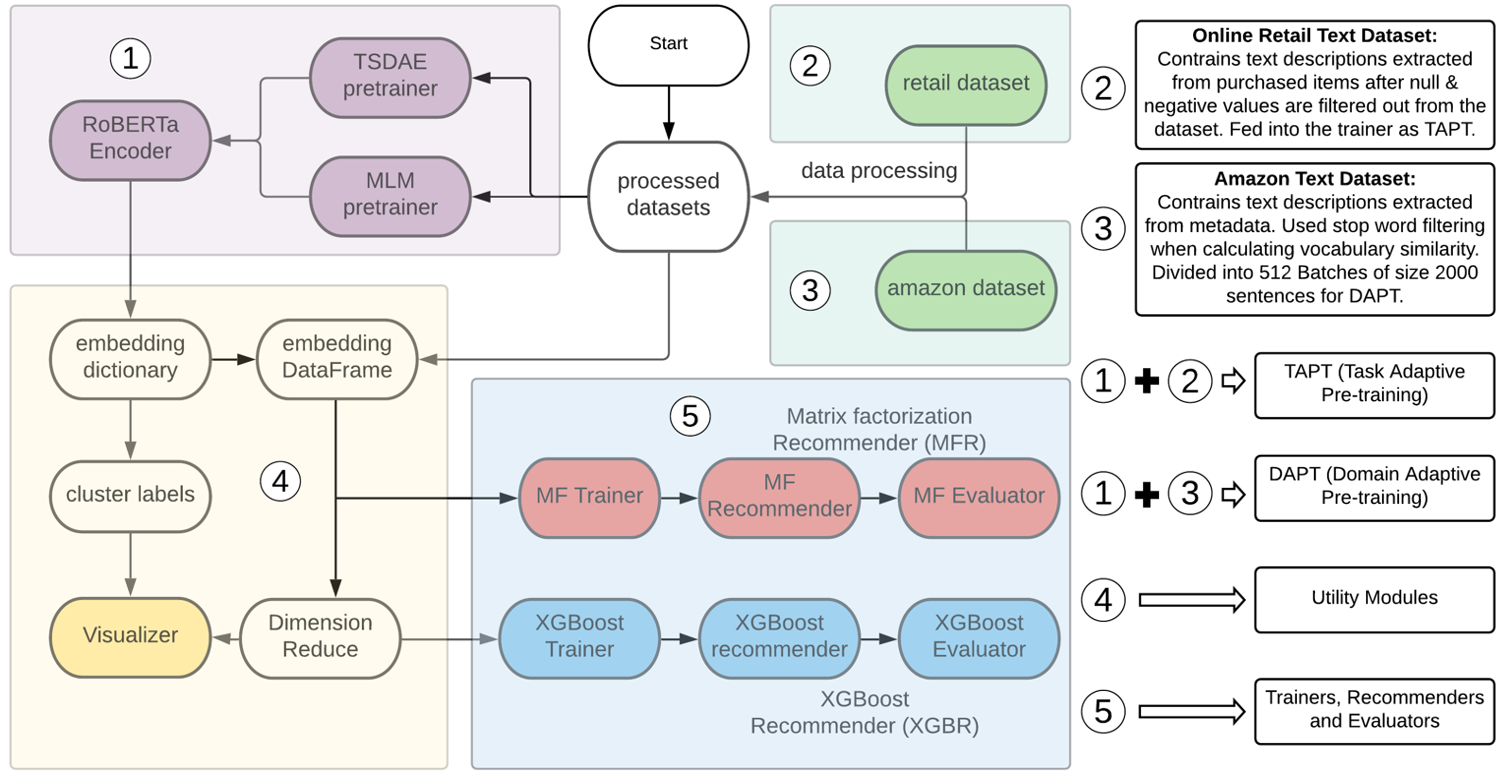}
	\caption[Close up of]
	{Flow chart of the data processing and training pipeline}
	\label{tab:figure4}
\end{figure*}

\section{Baseline Models}

We first build baseline models without using the text information of production description. Two main recommendation models are used in this paper:
\begin{itemize}
    \item Matrix Factorization Recommender 
    \item XGBoost Factorization Recommender
\end{itemize}
And for each approach, we have done investigation on the performance of three variants of those recommendation models. The details are listed in \autoref{tab:table1}.

\begin{table}[h!]
  \begin{center}
    \caption{Recommender models and variants}
    \label{tab:table1}
    \begin{tabular}{l|c|r} 
      \textbf{} & \textbf{Recommender 1} & \textbf{Recommender 2} \\
      variant & Matrix Factorization & XGBoost \\
      \hline
      1 & vanilla  & vanilla \\
      2 & generalized lower rank models & with random forest\\
      3 & implicit & with embedding features\\
    \end{tabular}
  \end{center}
\end{table}

\subsection{Matrix Factorization}
Standard matrix factorization-based collaborative filtering treats the entries of the user-item matrix as explicit user preferences given to items. Those algorithms assume explicit data, where the user has rated both things they like and dislike using some rating scale. They typically work by treating the missing data as an unknown, setting them to a constant (usually 0), and then the decomposed matrices are trained to reproduce the entire matrix, including the unknown values using stochastic gradient descent (SGD)\cite{Nguyen2013}.
It is, however, common in many real-world use cases that we only have access to implicit feedback (e.g. views, clicks, purchases, likes, shares). Essentially, instead of trying to model the matrix of ratings directly, implicit matrix factorization is used to treat the data as numbers representing the strength in observations of user actions (such as the number of clicks, or the cumulative duration someone spent viewing a movie). Those numbers are then related to the level of confidence in observed user preferences, rather than explicit ratings given to items. The model then tries to find latent factors that can be used to predict the expected preference of a user for an item\cite{Hu2008}.

Take the dataset used in this paper as an example, we can assume that a user buying an item means they like it, but we don't have the corresponding signal that a user doesn't like an item. Implicit data is usually more plentiful and easier to collect than explicit data - and even when you have the user give 5-star ratings the vast majority of those ratings are going to be positive only, so you need to account for implicit behaviour anyways. This means we can't just treat the missing data as unknowns, instead we have to treat a user not buying an item as being a signal that the user might not like that item. This presents a couple of challenges in learning a factorized representation.

The first challenge is in doing this factorization efficiently: by treating the unknowns as negatives, the naive implementation would look at every single entry in our input matrix. Since the dimensionality here is roughly 4K by 3K - there are over 12 million total entries to consider, compared to only 40K non-zero entries. The second problem is that we can't be certain that a user not listening to an artist actually means that they don't like it. There could be other reasons for the artist not being listened to, especially considering that we only have the top 50 most played artists for each user in the dataset.
To handle the case where we're not confident about our negative data, this approach learns a factorized matrix representation using different confidence levels on binary preferences: unseen items are treated as negative with a low confidence, where present items are treated as positive with a much higher confidence.

\subsection{XGBoost and Random Forest}
XGBoost builds decision tree one each time. Each new tree corrects errors which were made by previously trained decision trees\cite{Chen2016}. Since boosted trees are derived by optimizing an objective function, basically XGBoost can be used to solve almost all objective function that we can write gradient out. This includes things like ranking and Poisson regression, which RF is harder to achieve. XGBoost model, however, due to its strong modeling power, is more sensitive to over-fitting if the data is noisy. Training generally takes longer than RF because of the fact that trees are built sequentially. Additionally, XGBoost based models are usually harder to tune than RF. There are a few important hyper-parameters including: number of trees, depth of trees and learning rate, and the each tree built is generally shallow.\\

Random Forest (RF) trains each tree independently, using a random sample of the data\cite{Louppe2014}. This randomness helps to make the model more robust than a single decision tree. Thanks to that random forest is less likely to over-fit on the training data. The random forest dissimilarity helps to achieve high predictive accuracy for a high-dimensional problem with strongly correlated features, and has thus been used in a variety of applications. It also helps to reduce susceptibility against data that are highly noisy and that contains a lot of missing values e.g., some of the attributes are categorical or semi-continuous. The model tuning for RF is much easier than in case of XGBoost. In random forest we have two main parameters: number of features to be selected at each node and number of decision trees. The main limitation of the Random Forest algorithm is that a large number of trees can make the algorithm slow for real-time prediction. For data including categorical variables with different number of levels, random forests are biased in favor of those attributes with more levels.\\

XGBoost is normally used to train gradient-boosted decision trees and other gradient boosted models. Random Forests use the same model representation and inference, as gradient-boosted decision trees, but a different training algorithm. One can use XGBoost to train a standalone random forest or use random forest as a base model for gradient boosting.\\

In this paper, XGBoost is chosen as the recommender algorithm that models all the features extracted from the online retail dataset and provides recommendations for each user that are evaluated against ground truth purchases. RF algorithm is used as a feature selecter in an effort to minimize the influence of less useful features and underlying noises that could lead to over-fitting on the XGBoost model. 

\section{RoBERTa Fine-tuning}
\subsection{Corpus similarity}
\begin{figure}
	\centering
	\includegraphics[scale=0.6]{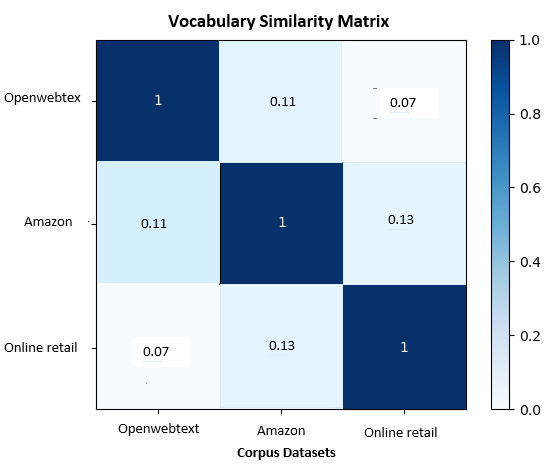}
	\caption[Vocabulary Similarity Matrix]
	{Vocabulary overlap (\%) between each language corpus. Vocabularies for each domain are created by considering the top 150 most frequent words in item descriptions sampled from each domain. A sample from openwebtext dataset is taken as a data source that is similar to RoBERTa’s pretraining corpus since the dataset used to train RoBERTa is no longer available.
    \label{tab:figure5}}
\end{figure}

\begin{figure}
	\centering
	\includegraphics[scale=0.45]{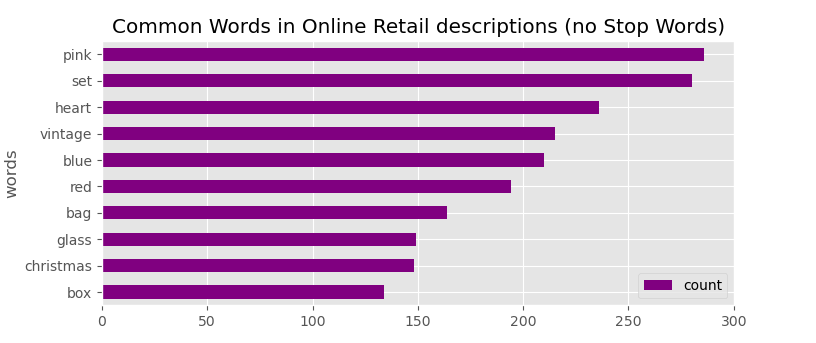}
	\caption[Retail Top 10 Words]
	{Top 10 common words found in descriptions of retail dataset with stop word filtered.}
	\label{tab:figure6}
\end{figure}

\begin{figure}
	\centering
	\includegraphics[scale=0.5]{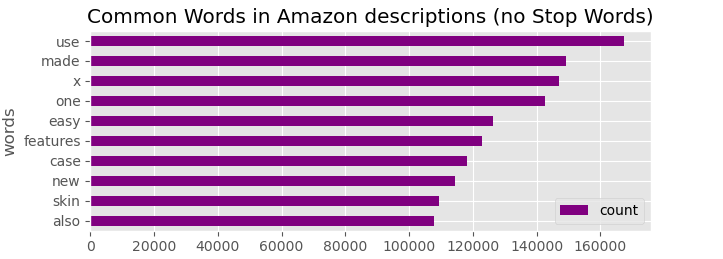}
	\caption[Amazon Top 10 Words]
	{Top 10 common words found in descriptions of amazon dataset with stop word filtered.}
    \label{tab:figure7}
\end{figure}

Two text datasets are used in this paper for RoBERTa fine-tuning. 
\begin{itemize}
    \item item descriptions in online retail dataset
    \item item descriptions in Amazon meta-data dataset
\end{itemize}
\autoref{tab:figure5} shows the word similarity between these two text datasets. \autoref{tab:figure6} and \autoref{tab:figure7}  shows the top 10 common words found in descriptions of online retail dataset and Amazon description dataset respectively after stop word filtering is applied.   
Note that stop word filtering has been used on both retail dataset and amazon dataset to get the word count. In comparison, descriptions in amazon dataset are much longer in character length, and contains less key information than in the case of retail dataset. This could potentially cause memory issues in GPU during training and add undesirable noises in the pre-trained result. To save memory, during the pretraining process, the sentences extracted from amazon dataset are divided into 512 different batches with a batch of size 2000, and the character limit is set to 50 in an effort to match the character length of the descriptions in retail dataset and to filter out less useful information in those descriptions.  

\subsection{Combined DAPT and TAPT}
Datasets curated to specific tasks of interest are typically a subset of textual information in a broader domain. Here we make the hypothesis that the online retail dataset is a narrowly defined subset of the broader e-commerce domain represented by amazon dataset due to its comparably smaller size. We adapt a similar pretraining scheme as used by Gururangan et al.\cite{Gururangan2020} that contains three different phases. We begin with regular RoBERTa pretraining, and then we use the online retail dataset and amazon review dataset for task adaptive pretraining (TAPT) and domain adaptive pretraining (DAPT) respectively. As suggested by Gururangan et al. \cite{Gururangan2020}, DAPT and TAPT might complement each other, depending on whether a spectrum of domains is defined around tasks at various levels of granularity.   \\
We investigate the effect of using both adaptation techniques together and observe the effect of manipulating the order of different pretraining phases. We begin with RoBERTa, and then apply one of DAPT or TAPT as a baseline. Then we explore two different orders of pretraining phases as shown in \autoref{tab:table5}. We show that combining domain- and task-adaptive pretraining achieves better performance on all tasks compared to their baseline counterparts. The results also demonstrate that applying TAPT after DAPT yields a better recommendation accuracy. 	

\section{Experimental Results}
Two main recommender models are used in the paper, Matrix factorization and XGBoost. Various variant approaches are taken to explore the key to better recommendation performance. The detailed list is shown in \autoref{tab:table1}.\\

We first establish baseline performance using vanilla Matrix Factorization and XGBoost models. Based on these, we add additional tweaks to see if there is any improvement and to better understand how to apply PLMs to those recommender systems. For visualizing and evaluating PLMs, we pass the sentences in our text datasets into those trained models and generate sentence embeddings of a fixed dimension of 728. Those high dimensional sentence embeddings are then passed through dimension reducing networks, where their dimension is reduced to either 10d or 2d depending on the settings. Two choices of dimension reduction methods are available, T-SNE and PCA. PCA provides more consistent result and faster running time\cite{Garber2015}, while T-SNE provides better clustering result while having slower running speed. T-SNE also does not work well on high dimensional vectors. In this paper, the embedding space is 2 dimensional, visualized through PCA and results are also generated using PCA. T-SNE are tested, and following \cite{VanDerMaaten2008}, we use PCA to reduce the dimension to 10d before we use T-SNE to further reduce it to 2d. However, T-SNE does not yield better result and the analysis regarding its performance is not relevant to our study in this paper.    \\

We use distilled RoBERTa as our baseline pretrained language model, with optional fine-tuning following two pretraining schemes introduced by Gururangan et al. \cite{Gururangan2020}, domain adaptive pretraining (DAPT) and task adaptive pretraining (TAPT). Our domain specific pretraining follows the settings prescribed for training RoBERTa. Due to the memory constraints, we did not choose to further pretrain RoBERTa model on openwebtext. Instead, we use RoBERTa as a baseline model, and investigate the effect of DAPT and TAPT using amazon description dataset and retail description dataset respectively. Two different kinds of pretraining method are used, Transformer-based Sequential Denoising Auto-Encoder (TSDAE) and masked language model (MLM). Studies have show that TSDAE is a strong pre-training method for learning sentence embeddings, significantly outperforming other approaches like MLM. It also provides better generalization and robustness for heterogeneous domains than the baseline RoBERTa model where domain knowledge is required. The pre-training is done in different orders to compare the effectiveness of different schemes. \\

\begin{table}
\centering
\begin{tabular}{ |p{1.8cm}||p{1.2cm}|p{1.2cm}|p{1.2cm}|p{1.2cm}| }
 \hline
 \multicolumn{5}{|c|}{Evaluation Result of XGBoost based algorithms} \\
 \hline
 & Accuracy  & Precision & Recall & F1 score\\
 \hline
 Baseline   & 71.2 (0.2)    & 70.1 (0.3)&   70.8 (0.4) & 70.4  (0.2)\\
 @ batch 100 &   72.3 (0.3)  & 71.4 (0.4)   & 69.9 (0.9) & 70.6 (0.3)\\
 @ batch 200 & 72.4 (0.2) & 72.2 (0.3)& 71.0 (0.6) & 71.6 (0.3)\\
 @ batch 300 & 73.0 (0.4) & 72.6 (0.6)&  72.2 (0.4) & 72.4 (0.3)\\
 @ batch 400 & 73.3 (0.2) & 72.8 (0.2) & 72.5 (0.3)& 72.6 (0.2) \\ 
 @ batch 500 & 73.1 (0.3) & 73.0 (0.2)& \textbf{72.8} (0.4)& 72.9 (0.3)\\
 @ final & \textbf{73.6} (0.6)& \textbf{73.4} (0.2)& 72.6 (0.4)& \textbf{73.0} (0.3)\\
 \hline
\end{tabular}
\caption{Prediction accuracy, precision, recall, and F1 score with standard deviation of baseline model (RoBERTa) calculated from 5 trial runs  and model that has been further pretrained using Amazon dataset using TSDAE. The above result is evaluated on the same amazon dataset using XGBoost except that the final model after trained with 512 batches is evaluated on retail dataset.}
\label{tab:table2}
\end{table}
\begin{table}
\centering
\begin{tabular}{ |p{1.8cm}||p{1.2cm}|p{1.2cm}|p{1.2cm}|p{1.2cm}| }
 \hline
 \multicolumn{5}{|c|}{Evaluation Result of XGBoost based algorithms} \\
 \hline
 & Accuracy  & Precision & Recall & F1 score\\
 \hline
 Baseline   & 67.4 (0.3)    & 68.2 (0.2)&   67.0 (0.3) & 64.8 (0.3) \\
 @ batch 100 &   72.2 (0.3)  & 71.6 (0.2)   & 69 (0.2) & 70.3 (0.3)\\
 @ batch 200 & 72.4 (0.2) & 72.5 (0.3)& 71.0 (0.1) & 71.7 (0.4)\\
 @ batch 300 & 73.0 (0.2) & 72.8 (0.2)&  72.2 (0.3) & 72.5 (0.1)\\
 @ batch 400 & 73.3 (0.3) & 73.0 (0.3) & 72.5 (0.5)& 72.7 (0.4) \\ 
 @ batch 500 & 74.0 (0.1) & 73.6 (0.3) & \textbf{73.2} (0.2)& 73.4 (0.2)\\
 @ final & \textbf{74.2} (0.2) & \textbf{74.5} (0.4)& 72.8 (0.3)& \textbf{73.6} (0.2)\\
 \hline
\end{tabular}
\caption{Prediction accuracy, precision, recall, and F1 score with standard deviation of baseline model (RoBERTa) calculated from 5 trial runs  and model that has been further pretrained using Amazon dataset using MLM. The above results are evaluated on the same amazon dataset using XGBoost except that the final model after trained with 512 batches is evaluated on retail dataset.}
\label{tab:table3}
\end{table}

\begin{figure}[h]
	\centering
	\includegraphics[scale=0.6]{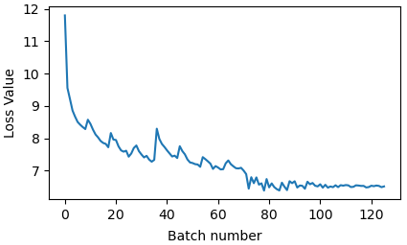}
	\caption[Visualization of embedding space distribution of items recommended by matrix factorization model and the ground truths]
	{Loss value vs batch number for MLM pretraining graph evaluated on amazon dataset}
	\label{tab:figure8}
\end{figure}

First, we investigate the effectiveness of DAPT and TAPT individually. Amazon sentences are divided into 512 batches of size 2000, with maximum characters allowed in each sentence set to 50. Then those batches are fed into two different pretraining frameworks, TSDAE and MLM respectively. Prediction accuracy, precision, recall and F1 score are used as evaluation metrics and model performance is evaluated every 100 batch and at the end. The results are listed in \autoref{tab:table2} and \autoref{tab:table3}. The performance of XGB recommender increases with training for all evaluation metrics and the speed becomes slower with the increase of batch number. This can also be supported by the training loss graph shown in \autoref{tab:figure8}. Training loss drops fast at first 100 batches and then becomes stable after that.\\ 

Then we explore different orders of combining TSDAE and MLM pretraining, with different combinations of DAPT and TAPT, in comparison to the baseline performance we just established above. The result is listed in \autoref{tab:table4}. \\

\begin{table}[ht!]
\centering
\begin{tabular}{ |p{2.2cm}||p{1.1cm}|p{1.1cm}|p{1.1cm}|p{1.1cm}| }
 \hline
 \multicolumn{5}{|c|}{Evaluation Result of XGBoost based algorithms} \\
 \hline
 & Accuracy  & Precision & Recall & F1 score\\
 \hline
 TSDAE @amazon  & 71.2 (0.2)    & 70.1 (0.3)&   70.8 (0.4) & 70.4 (0.2) \\
 MLM @amazon &   72.3 (0.6)  & 71.1 (0.4)   & 69.8 (0.4) & 70.4 (0.2)\\
 T\textrightarrow M @amazon & 72.4 (0.3) & 72.2 (0.2)& 71.4 (0.3) & 71.8 (0.3)\\
 M\textrightarrow T @amazon & 73.0 (0.1) & 72.6 (0.3)&  72.2 (0.2) & 72.4 (0.4)\\
 T\textrightarrow M @retail & 73.3 (0.2) & 72.8 (0.4) & 72.5 (0.6)& 72.6 (0.1) \\ 
 M\textrightarrow T @retail & 73.1 (0.4) & 73.0 (0.2) & 72.8 (0.4)& 72.9 (0.3)\\
 MLM @D\textrightarrow T & 73.6 (0.2)& 73.4 (0.4)& 72.6 (0.2)& 73.0 (0.2)\\
 TSDAE @D\textrightarrow T & 73.5 (0.3) & 73.2 (0.2)& 72.6 (0.1)& 72.9 (0.1)\\
 T\textrightarrow M @D\textrightarrow T  & 74.4 (0.2) & 74.3 (0.2)& \textbf{73.1} (0.2)& 73.7 (0.2)\\
 M\textrightarrow T @D\textrightarrow T & \textbf{74.8} (0.2)& \textbf{74.5} (0.3)& 73.0 (0.4)& \textbf{73.7} (0.5)\\
 \hline
\end{tabular}
\caption{Prediction accuracy, precision, recall, and F1 score with standard deviation calculated from 5 trial runs. T\textrightarrow M stands for TSDAE\textrightarrow MLM, D\textrightarrow T stands for DAPT\textrightarrow TAPT.}
\label{tab:table4}
\end{table}

We also study the influence of various features used in generating recommendations by XGBoost model. Random Forest algorithm is used as an optional feature selection method that filters out features with importance less than 0.04 in an attempt to reduce the chance of XGBoost model overfitting on less useful features. Different features are added in a paralleled manner except those with a * after the feature name. The details are listed in \autoref{tab:table5}.\\

\begin{table}[htbp]
\centering
\begin{tabular}{ |p{1.8cm}||p{1.2cm}|p{1.2cm}|p{1.1cm}|p{1.1cm}| }
 \hline
 \multicolumn{5}{|c|}{Evaluation Result of XGB based algorithms} \\
 \hline
 & Accuracy  &Precision &Recall &F1 score\\
 \hline
 Vanilla XGB   & 54.2 (0.3)    & 52.3 (0.2)&   49.5 (0.2) & 67 (0.1) \\
 +RandomForest &   56.1 (0.4)  & 55.4 (0.2)   & 51.9 (0.3) & 56.4 (0.6)\\
 +cluster label & 60.2 (0.3) & 61.2 (0.3)& 55.1 (0.1) & 58.7 (0.3)\\
 +10d embed & 73.0 (0.2) & 72.8 (0.5)&  72 (0.3) & 73 (0.2)\\
 +2d embed & 73.3 (0.3) & 73.0 (0.4) & 72.5 (0.2)& 73 (0.5) \\ 
 best TAPT  & 74.0 (1.2) & 73.6 (0.8) & 73.2 (0.9)& 73.6 (0.6)\\
 +stock code & 74.2 (0.3) & 74.5 (0.4)& 72.8 (0.5)& 73.9 (0.2)\\
 +invoice no. & 74.2 (0.6) & 74.5 (0.3)& 72.8 (0.3)& 73.9 (0.6)\\
 DATP\textrightarrow TAPT & 74.2 (0.3) & 74.5 (0.5)& 72.8 (0.4)& 73.9 (0.4)\\
 \hline
\end{tabular}
\caption{Prediction accuracy, precision, recall, and F1 score of the final XGBoost model with standard deviation (calculated from a total of 5 trial runs) using different number of features. The result is evaluated on retail dataset.}
\label{tab:table5}
\end{table}

Since the number of purchased items are labeled as 0, 1, 2, 3, 4, we also want to check the performance of the model on each label, namely, on each range of purchase numbers. The result is shown in \autoref{tab:table6}. \\  

\begin{table}
\centering
\begin{tabular}{ |p{0.8cm}||p{1.2cm}|p{1.2cm}|p{1.6cm}|p{1.6cm}| }
 \hline
 \multicolumn{5}{|c|}{Evaluation Result of XGB based algorithms} \\
 \hline
 Label & TSDAE  & MLM & T\textrightarrow MLM & MLM\textrightarrow T\\
 \hline
0  & 75.3 (0.2)    & 77.6 (0.4)&   80.2 (0.4) & 82.0 (0.2) \\
1 &   50.4 (0.2)  & 52.4 (0.2)   & 53.9 (0.3) & 55.4 (0.2)\\
2 & 58.7 (0.2) & 60.4 (0.2)& 60.8 (0.3) & 62.6 (0.1)\\
3 & 77.8 (0.3) & 78.1 (0.3)&  78.4 (0.4) & 79.2 (0.3)\\
4 & 79.2 (0.2) & 79.4 (0.1)& 80.0 (0.2)& 81.4 (0.3)\\
 \hline
\end{tabular}
\caption{Maximum prediction accuracy achieved by DAPT followed by TAPT with different pretraining strategies, averaging 3 trial runs with standard deviation.}
\label{tab:table6}
\end{table}

\begin{table}
\centering
\begin{tabular}{ |p{2.1cm}||p{1.4cm}|p{1.1cm}|p{1.1cm}|p{1.1cm}| }
 \hline
 \multicolumn{5}{|c|}{Evaluation Result of MF based algorithms} \\
 \hline
 Label & Precision@5  & Recall@5 & F1 score & MAP\\
 \hline
vanilla MF  & 23.4 (0.3)  & 3.4 (0.1) & 5.9 (0.2) & 54.6 (0.1) \\
GLRM & 26.4 (0.2)  & 4.8 (1.2)  &  8.1 (1.4) & 57.4 (0.9)\\
Implicit MF & 33.4 (0.5) & 6.2 (1.4)& 10.5 (1.1) & 58.6 (2.4)\\
+cluster label & 35.6 (0.8) & 6.4 (0.7)&  10.8 (0.8) & 58.9 (0.8)\\
+embed similarity & \textbf{36.2}  (1.0)& \textbf{6.6} (1.2)& \textbf{11.2} (1.3) & \textbf{60.2} (2.3)\\
 \hline
\end{tabular}
\caption{Precision, recall, F1 score and MAP @5 evaluated on retail dataset using matrix factorization with different matrix factorization algorithms. * means the dataset has been filtered as following: any user that has purchased less than 5 items and any item that has been purchased less than 5 times are removed from the dataset for training and evaluation.}
\label{tab:table7}
\end{table}

\begin{table}
\centering
\begin{tabular}{ |p{2.0cm}||p{1.4cm}|p{1cm}|p{1.1cm}|p{1.1cm}| }
 \hline
 \multicolumn{5}{|c|}{Evaluation Result of MF based algorithms} \\
 \hline
 Label & Precision@5  & Recall@5 & F1 score & MAP\\
 \hline
TSDAE @retail  & 35.8 (0.7)    & 6.4 (0.7)&   10.9 (0.7) & 59.2 (0.1) \\
MLM @retail &   36.3 (0.3)  & 6.6  (0.6)  & 11.2 (0.9) & 60.6 (0.6)\\
T\textrightarrow M @retail & 36.3 (0.8) & 6.6 (1.2) & 11.2 (1.4) & \textbf{60.4} (0.9)\\
M\textrightarrow T @retail & \textbf{36.4} (0.3) & \textbf{6.7} (0.5)&  \textbf{11.2} (0.6) & 60.2 (0.4)\\
 \hline
\end{tabular}
\caption{Precision, recall, F1 score and MAP @5 evaluated on retail dataset using matrix factorization with different strategies of pretraining listed in first column. The matrix factorization method used here is implicit matrix factorization with cosine similarity between sentence embeddings generated by pretrained TAPT model. }
\label{tab:table8}
\end{table}

\begin{table*}[t]
\centering
\begin{tabular}{ |p{2.6cm}||p{1cm}|p{1cm}|p{1cm}|p{1cm}|p{1cm}|p{1cm}|p{1cm}|p{1cm}|p{1cm}|p{1cm}|}
 \hline
 \multicolumn{11}{|c|}{Evaluation Result of MF based algorithms} \\
 \hline
 & item 1 & item 2 & item 3&item 4&item 5&item 6&item 7&item 8&item 9& item 10\\
 \hline
recommendation & 1605 & 266 & 1931 &1930 &749 &620 & 751 &1937 &2609 & 2612 \\
cluster label & 7 & 5& 5& 6 & 7 & 4 & 5 & 5 & 7 & 7\\
ground truth & 170 & 171 & 1203 &1205& 1253 &1410& 1482& 1483 & 1484 & N/A\\
cluster label & 7 & 11 & 1 &1&2&9&4 & 4 & 4 & N/A\\
predict correctness & \xmark & \xmark & \xmark & \xmark& \xmark& \xmark& \xmark& \xmark& \xmark & N/A\\
label correctness & \checkmark & \xmark & \xmark & \xmark& \xmark& \xmark& \xmark& \xmark& \xmark & N/A\\
 \hline
\end{tabular}
\caption{Recommendation and cluster label prediction with evaluation result for a user example that the model does not perform well on.}
\label{tab:table9}
\end{table*}
As shown in \autoref{tab:table7}, recall @5 is quite low in comparison to precision @5. This is understandable because of the characteristics of the retail dataset used in this paper. Lots of the customers of this online retailer are also retailers from other industry, and they usually purchase a lot of items of different types at once. As a fact, there are costumers who bought more than 40 items. Even if the 5 recommendations made by the model are all correct, that would only give us a 5/40 = 12.5\% recall @5. Therefore, recall @5 is not a very good metric here to evaluate the performance of the matrix factorization model as the number of recommendations made here is typically much lower than the amount of items that got purchased by the customers. A better metric to look at here would be precision based, namely, f1 score or MAP(mean average precision).\\

We speculate that there is no strong interrelationship between different customers who have done purchases at this online retail. Customers tend to buy items that belong to their industry. For example, furniture retailers like to buy crews and other related components, and normally in a large number. That does not relate well with another user, say an individual buyer. The customer\_id feature whose introduction leads to a boost in model performance supports this speculation. \\
\begin{figure}[h]
	\centering
	\includegraphics[scale=0.42]{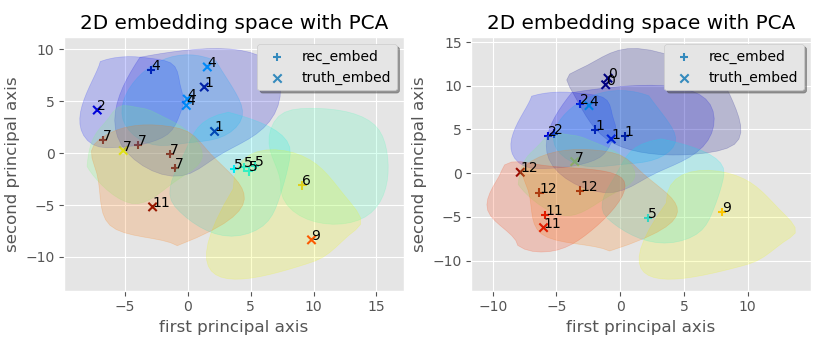}
	\caption[Visualization of embedding space distribution of items recommended by matrix factorization model and the ground truths]
	{embedding space visualization for user 0 (left) and user 1 (right)}
	\label{tab:figure9}
\end{figure}

\begin{figure}[h]
    \begin{subfigure}[b]{0.24\textwidth}
      \includegraphics[width=\textwidth]{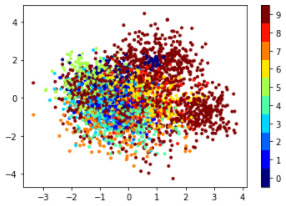}
      \caption{PCA}
    \end{subfigure}
    \begin{subfigure}[b]{0.24\textwidth}
      \includegraphics[width=\textwidth]{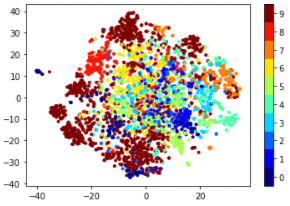}
      \caption{T-SNE}
    \end{subfigure}
    \caption{embedding space generated using PCA (left) and T-SNE (right), with each dot representing the embedding vector of a specific product description}
    \label{tab:figure10}
\end{figure}
 
Two user examples on which the model does not recommend items accurately are chosen here for visualization to better understand what happened. The examples are visualized through PCA and shown in \autoref{tab:figure9}. We evaluate them based on two criteria, correctness on the predicted cluster label and predicted stock code of the item. For the left example, the details of which are shown in \autoref{tab:table9}, only 1 out of 10 cluster label is predicted correctly, and none of the stock codes. From \autoref{tab:figure9}, we observe that predicted items are mostly clustered around cluster 7 and cluster 5, while the ground truth items spread around them. This means that the recommender has picked up some patterns, and tries to focus on some dimensions of the latent space that it considers to be important. However, due to lack of modeling power, it could not fully exploit the rich information contained in the sentence embeddings. Similar pattern can also be observed from the right figure in \autoref{tab:figure9}. As a reference, \autoref{tab:figure10} show the difference between the distributions of embedding vectors visualized in 2d through PCA (left) and T-SNE (right) respectively. The result obtained by PCA is more spread out while the result obtained by T-SNE is more tightly clustered. However, due to the high dimensional nature of sentence embeddings, it remains difficult to accurately interpret the result in 2D, which is subject to visualization methods.\\

\section{Discussion and Future Work}
Experiments have shown that MF-based recommenders do not work well with sentence embeddings when they are taken directly as the recommendation algorithms. Attempts are made to combine sentence embeddings generated by PLMs as item representations with item features generated by MF model. The approaches taken include: 
\begin{itemize}
    \item combining top ranked recommendation results from MF recommender and PLM recommender that ranks items based on cosine similarity of their sentence embeddings. 
    \item taking each entry of the UI matrix as a weighted sum of the dot product of user vector and item vector from MF model and the similarity score from the PLM  
    \item using sentence embeddings of items purchased by other users who are considered to be similar user by MF model as auxiliary information to facilitate the MF recommender
\end{itemize}
However, none of the above-mentioned approaches lead to any observable improvement in recommendation performance. This shows that, due to the high dimensional nature of sentence embeddings and the characteristics of this Online Retail Dataset, MF is not the ideal model to be used directly as the recommendation algorithm. A model that better leverages the rich information contained within those sentence embeddings is needed. However, as suggested by LadaBERT\cite{Mao2021}, when matrix factorization is used as a model compression technique alongside with weight pruning and knowledge distillation, BERT model can be made less memory and data intensive. It also helps the model to perform better with significantly fewer training overheads. It is not covered in this paper and can be an interesting future direction.

We find that the introduction of transformer based pretrained language embedding model can boost the performance of traditional recommendation algorithms, especially on XGBoost recommender. Through carefully designed fune-tuning on RoBERTa model using TSDAE and MLM, the recommendation accuracy can be further improved. Empirical data shows that MLM followed by TSDAE achieves a better performance increase, as we speculate that MLM pretraining provides a rudimentary domain specific knowledge to the model and TSDAE further increases generalization and robustness across heterogeneous domains. 

The vocabulary similarity between amazon description dataset and retail description dataset is comparably low, and this might compromise the effectiveness of domain specific pretraining. Also due to GPU memory constraints, the datasets used in this paper are small scaled in comparison to language corpus that are used to train RoBERTa. A dataset that is larger in scale and more structurally similar to the online retail dataset can be used in the future to investigate the full potential of this pretraining method.   

There are more state-of-art deep learning based recommendation algorithms that could be used as the recommendation model. In this paper we choose to focus on an investigation on the effectiveness of the introduction of pre-trained language models into recommender systems using matrix factorization algorithm or XGBoost algorithm. Exploring the benefits of PLMs on deep learning based recommender systems will be an interesting future direction. 

As suggested and experimented by Zhang et al.\cite{Zhang2020}, Adaptive Hybrid Masking (AHM) extends MLM by introducing a new masking
strategy. Specifically, it sets two different modes, i.e., word masking mode and phrase masking mode. The former randomly masks separate words while the latter masks domain phrases. Moreover, it can adaptively switch between the two modes based on feedback losses, enabling the model to capture word-level and phrase-level knowledge progressively. Compared with MLM, it has a higher potential to increase the model performance if it is chosen to be the fine-tuning method. Additionally, like AHM, new advanced pre-training methods can be used instead of the two that are used in this paper in the future.    

\section{Conclusion}
Recent empirical improvements due to the incorporation of PLMs and thoroughly designed fine-tuning strategies into traditional recommender systems have been shown that rich, staged, and unsupervised pre-training is a vital and indispensable part of boosting model performance in the newly thriving field of e-commerce. Our major contribution is to investigate various methods of combining PLMs and fine-tuning with MF and XGBoost recommendation algorithms, and propose a fine-tuning scheme that allows the traditional model to outperform its vanilla counterpart respectively. 

\clearpage
\bibliographystyle{apalike}
\bibliography{references}
\end{document}